\documentclass[9pt,conference]{IEEEtran}
\IEEEoverridecommandlockouts
\usepackage{cite}
\usepackage{amsmath,amssymb,amsfonts}
\usepackage{algorithmic}
\usepackage{graphicx}
\usepackage{textcomp}
\usepackage{xcolor}
\usepackage{amsmath,amssymb,amsfonts}
\usepackage{bbm}
\usepackage{dsfont}
\usepackage{booktabs}
\usepackage{multirow}
\usepackage{makecell}
\usepackage{pifont}
\newcommand{\cmark}{\ding{51}}
\newcommand{\xmark}{\ding{55}}
\usepackage[linesnumbered,ruled,vlined]{algorithm2e}
\usepackage{wrapfig}

\def\BibTeX{{\rm B\kern-.05em{\sc i\kern-.025em b}\kern-.08em
    T\kern-.1667em\lower.7ex\hbox{E}\kern-.125emX}}
\begin{document}

\title{Reinforced Domain Selection for Continuous \\Domain Adaptation}

\author{%
  \IEEEauthorblockN{%
    Hanbing Liu, 
    Huaze Tang,
    Yanru Wu,
    Yang Li\IEEEauthorrefmark{2} and
    Xiao-Ping Zhang
  }%

  \IEEEauthorblockA{\textit{Shenzhen Key Laboratory of Ubiquitous Data Enabling, Tsinghua Shenzhen International Graduate School, Tsinghua University}}
  \IEEEauthorblockA{ \{liuhb24,thz21,wu-yr21\}@mails.tsinghua.edu.cn, yangli@sz.tsinghua.edu.cn, xpzhang@ieee.org}
  \thanks{\IEEEauthorrefmark{2} Corresponding authoress.}
}

\maketitle

\begin{abstract}
Continuous Domain Adaptation (CDA) effectively bridges significant domain shifts by progressively adapting from the source domain through intermediate domains to the target domain. However, selecting intermediate domains without explicit metadata remains a substantial challenge that has not been extensively explored in existing studies. To tackle this issue, we propose a novel framework that combines reinforcement learning with feature disentanglement to conduct domain path selection in an unsupervised CDA setting. Our approach introduces an innovative unsupervised reward mechanism that leverages the distances between latent domain embeddings to facilitate the identification of optimal transfer paths. Furthermore, by disentangling features, our method facilitates the calculation of unsupervised rewards using domain-specific features and promotes domain adaptation by aligning domain-invariant features. This integrated strategy is designed to simultaneously optimize transfer paths and target task performance, enhancing the effectiveness of domain adaptation processes. Extensive empirical evaluations on datasets such as \textit{Rotated MNIST} and \textit{ADNI} demonstrate substantial improvements in prediction accuracy and domain selection efficiency, establishing our method's superiority over traditional CDA approaches.

\end{abstract}


\section{Introduction}

Domain shift is a common challenge in many real-life applications \cite{chen2022self}. Continuous Domain Adaptation (CDA) strategically mitigates the challenge of significant domain shifts by seamlessly transferring from the source domain through various intermediate domains to the target domain \cite{zhang2021adaptive, pan2009survey}. Traditional CDA methods such as self-training, pseudo-labeling, adversarial algorithms, and optimal transport have advanced significantly \cite{moon2020multi, xu2022delving, wang2020continuously, liu2024enhancing, zhou2022active, kumar2020understanding, hoffman2014continuous, liang2019distant, liang2020we, bitarafan2016incremental, tzinis2022continual}. However, the dynamic selection of intermediate domains without explicit metadata remains a complex problem \cite{ortiz2019cdot}. To address this issue, novel methodologies have been proposed to enhance domain sorting and minimize errors. For instance, \cite{chen2021gradual} introduces a progressive domain discriminator in their work, enabling the generation of domain sequences without pre-defined domain indexes. Furthermore, \cite{liu2024enhancing} introduced a Wasserstein-based transfer curriculum that strategically sorts intermediate domains using Wasserstein distance, reducing cumulative errors through a multi-path strategy. Nonetheless, these methods mainly focus on sorting available domains. 
The resulting sequence of domains, a.k.a. transfer path, is not guaranteed to be optimal: sparser path would lead to larger gaps between subsequent domains, prone to negative transfer, while denser path requires longer training time and is more likely to accumulate errors.  
Hence, it is necessary to incorporate dynamic transfer path learning during the adaptation process  
to incorporates only the essential domains.

\begin{figure*}[ht]
	\begin{center}
		\centerline{\includegraphics[width=1.8\columnwidth]{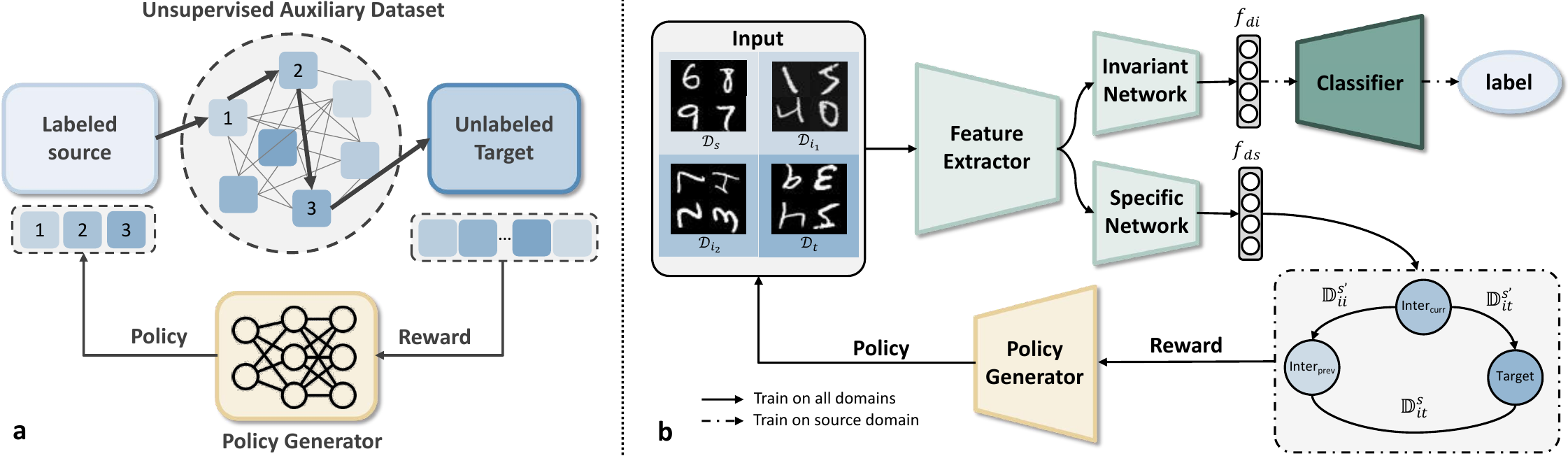}}
		\caption{\textbf{Overview and framework of our method.} (a) Overview of Continual Domain Adaptation using Reinforcement Learning. Our approach employs a policy generator to devise strategies for intermediate domains, thus establishing an optimal transfer path. (b) Framework of the Proposed Method. Input from the source, target, and intermediate domains is processed by a feature extractor to derive common features. Subsequently, a dual-network system isolates domain-invariant and domain-specific features. The domain-specific features from various domains are then evaluated based on their distances to calculate rewards, which assist the policy generator in formulating policies for each intermediate domain.}
		\label{fig:overview}
	\end{center}
 \vspace{-0.5cm}
\end{figure*}

Learning the optimal path among multiple intermediate domains poses a significant combinatorial optimization challenge due to its dynamic nature \cite{xu2018reinforced}. In response, we propose a novel Reinforcement Learning (RL) strategy for dynamically selecting intermediate domains during the CDA process as illustrated in Figure \ref{fig:overview}(a). Existing methods combining RL with continual domain adaptation, such as \cite{huang2022curriculum, gao2022efficient, xu2018reinforced, chen2018policy}, typically focus on dynamically expanding network models, which significantly increases computational demands. Diverging from these approaches, we are inspired by \cite{liu2019reinforced}, which utilized policy gradient methods for training data selection in supervised settings. However, their reliance on target labels for reward calculation is not viable in unsupervised settings. Additionally, the omission of domain continuity limits its applicability to the CDA challenge. 

Therefore, we propose a surrogate reward function based on the distance between feature distributions of different domains to learn the optimal domain selection policy without supervision. Furthermore, features are inherently high-dimensional, which requires longer training time for the system to converge to an optimal policy \cite{mowakeaa2021kernearl}. We leverage the finding in CDA that task-related information is typically invariant to domain shifts \cite{lao2020continuous, peng2019domain}. By disentangling features into domain-invariant and domain-specific components, our approach not only learns a domain-agnostic model but also enhances the accuracy of domain shift estimations through the application of low-dimensional latent domain embeddings. To the best of our knowledge, this study is the first to integrate reinforcement learning with feature disentanglement to tackle the domain selection challenge in CDA scenario. Extensive evaluations across handwritten digits and medical image classification datasets demonstrate our approach's superiority, enhancing prediction accuracy and path selection strategy over traditional CDA methods. The key contributions of this study are outlined as follows:

\begin{itemize} 
\item[1)] \textbf{RL-based intermediate domain selection:} We formulate the problem of dynamically selecting intermediate domains using RL with feature disentanglement to optimize the transfer path, focusing on simultaneous optimization of the transfer path and prediction outcomes. 
\item[2)] \textbf{Novel reward mechanism:} Our novel domain selection policy employs an unsupervised reward mechanism based on the distance between latent domain embeddings, enhancing strategy precision. 
\item[3)] \textbf{Disentangled domain embedding:} We separate domain-specific from domain-invariant features to improve the extraction of transferable features for domain adaptation, while enabling more precise domain shift estimations through low-dimensional embeddings. 
\end{itemize}

\section{Methodology}

\subsection{Problem Definition}

This research investigates a CDA setting involving multiple unlabeled auxiliary domains accompanying a labeled source domain. The domain indices are defined as $\mathcal{G} = \mathcal{G}_s \cup \mathcal{G}_t \cup \mathcal{G}_i$, categorizing each domain into source ($\mathcal{G}_s$), target ($\mathcal{G}_t$), and intermediate ($\mathcal{G}_i$). The source domain $\mathcal{D}_s$ consists of labeled tuples $(\boldsymbol{x^{s}_j}, y^{s}_j, g^{s}_j)$, where $\boldsymbol{x^{s}_j}$ represents feature vectors, $y^{s}_j$ are the labels, and $g^{s}_j$ indicates the domain index for each $j=1, \dots, n_s$. The unlabeled target domain, denoted as $\mathcal{D}_t$, comprises tuples $(\boldsymbol{x^{t}_{j}}, g^{t}_{j})$ for $j = 1, \dots, n_t$. Similarly, the unlabeled intermediate domain $\mathcal{D}_{i_k}$ includes tuples $(\boldsymbol{x^{i}_{j}}, g^{i}_{j})$ for $j = 1, \dots, n_{i_k}$ and $k = 1, \dots, K$, which means there are $K$ intermediate domains. $\boldsymbol{x^{t}_{j}}$ and $\boldsymbol{x^{i}_{j}}$ are represented as feature vectors, indexed by $g^{t}_{j} \in \mathcal{G}_t$ and $g^{i}_{j} \in \mathcal{G}_i$. This reinforced domain selection problem in CDA can be formulated by its transfer path $h = (\mathcal{G}_{h_1}, \mathcal{G}_{h_2}, \dots, \mathcal{G}_{h_L})$ where $L\leq K$ and $\{h_l\}_{l=1}^{L}$ is the domain id of the $l$-th intermediate domain in the path. The primary objective of this research is to derive an optimal path $\hat{h}$, such that by transferring along $\hat{h}$ during training, we accurately predict the labels $(y^{t}_{j})^{n_t}_{j=1}$ for the feature vectors in the unlabeled target domain $\mathcal{D}_t$, utilizing both the labeled data from the source domain and the structural insights gleaned from the intermediate domains.

\subsection{Method Framework} 

We present a novel approach that integrates reinforcement learning with feature disentanglement to address the challenges of domain selection in CDA, as depicted in Figure \ref{fig:overview}(b). Our methodology utilizes a dataset consisting of a labeled source domain, multiple unlabeled intermediate domains, and a target domain. A feature extractor \(F\) isolates common features across these domains, which are subsequently processed by an invariant network \(I\) and a specific network \(S\). The invariant network extracts domain-invariant features \(f_{di}\), while the specific network focuses on domain-specific features \(f_{ds}\), with mutual information ensuring the independence of these components. Subsequently, \(f_{ds}\) from intermediate domains feed into a policy generation network \(P\), which formulates a selection strategy via a policy gradient algorithm based on the discrepancies between the domains. During inference, the trained networks \(\hat{F}\), \(\hat{I}\), and classifier \(\hat{C}\) collaborate to predict labels in the target domain.

\subsection{Feature Disentanglement}
\label{sec:disentanglement}

To enhance feature alignment and decomposition, we employ a feature extractor $F$, along with an Invariant network $I$ and a Specific network $S$. Initially, all domains are input into the feature extractor to derive common features. These are subsequently handled by networks $I$ and $S$ to extract domain-invariant and domain-specific features, respectively, ensuring targeted feature isolation for subsequent analysis. 
The features are categorized into two types: domain-invariant feature \( f_{di} \) and domain-specific feature \( f_{ds} \). These features are further divided into three components for each domain type: \( f_{di}^s \), \( f_{di}^i \), and \( f_{di}^t \) which represent the domain-invariant features of the source, intermediate, and target domains, respectively. Similarly, \( f_{ds}^s \), \( f_{ds}^i \), and \( f_{ds}^t \) correspond to the domain-specific features. The features $f_{di}$ and $f_{ds}$ are derived by modulating Mutual Information (MI) \cite{kraskov2004estimating}, a fundamental metric quantifying the dependency between two random variables. Given the substantial computational complexity of calculating MI directly, we utilize the Mutual Information Neural Estimator (MINE) \cite{belghazi2018mutual}, which provides a practical approach to estimate MI from $n$ i.i.d. samples using a neural network $T_\theta$. In our neural network model, the lower bound of mutual information estimation is implemented as
\begin{equation} I(X ; Z) = \frac{1}{n} \sum_{i=1}^n T_\theta(\boldsymbol{x}^{(i)}, \boldsymbol{z}^{(i)}) - \log \left(\frac{1}{n} \sum_{i=1}^n e^{T_\theta(\boldsymbol{x}^{(i)}, \overline{\boldsymbol{z}}^{(i)})}\right),
\label{eq:mi}
\end{equation}
where $(\boldsymbol{x}^{(i)}, \boldsymbol{z}^{(i)})$ represent samples from the joint distribution and $\overline{\boldsymbol{z}}^{(i)}$ are sampled from the marginal distribution. This neural network-based methodology offers a scalable solution for estimating MI in extensive datasets.

The loss functions for the invariant and specific networks, $\mathcal{L}_{mi}$ and $\mathcal{L}_{ms}$ respectively, are defined as follows:

\begin{equation}
        \mathcal{L}_{mi} = -[I(f_{di}^s; f_{di}^i) +I(f_{di}^s; f_{di}^t)+I(f_{di}^i; f_{di}^t)],
\label{eq:lmi}
\end{equation}
\begin{equation}
        \mathcal{L}_{ms}  = I(f_{ds}^s; f_{ds}^i)+I(f_{ds}^s; f_{ds}^t)+I(f_{ds}^i; f_{ds}^t),
\label{eq:lms}
\end{equation}
where $I$ denotes the mutual information estimation as defined in Equation \ref{eq:mi}. The invariant network strives to maximize the pairwise mutual information, while the specific network aims to minimize it. Subsequently, the unified domain-invariant feature \( f_{di}^s \) from the source domain is classified by the classifier $C$, utilizing cross-entropy loss $\mathcal{L}_{ce}$ for supervised learning:
\begin{equation}
\mathcal{L}_{ce}=-\mathbb{E}\left[ \sum_{m=1}^{M} \mathbbm{1}\left[m=y^s\right] \log \left(C(f_{di}^s)\right)\right],
\label{eq:ce}
\end{equation}
where $M$ denotes the number of categories in a classification problem. By isolating both unified and distinctive features, we improve the efficiency of transfer processes and support the development of policy generation.

\begin{algorithm}[ht]
\DontPrintSemicolon
  \KwIn{Source domain $\mathcal{D}_s$, Target domain $\mathcal{D}_t$, Intermediate domains $\mathcal{D}_{i_k}, k = 1, \dots, K$, transfer path $h$, Feature Extractor $F$, Invariant network $I$, Specific network $S$, and Policy Generator $P$.}
  
  \KwOut{Well-trained $\hat{F}, \hat{I}, \hat{S}, and \hat{P}$.}
  
  \BlankLine

\textbf{Initialize: } $F, I, S, P \sim$ Xavier initialization

\For{$e \leftarrow 1, ...$}{
\For{$k \leftarrow 1, ..., K$}{
Shuffle intermediate domain set \\
\For{$\mathcal{D}_{i_k}$ in intermediate domain set}{
\textbf{Feature Disentanglement} \\
update $F, I, C$ using Equation \ref{eq:lmi} $+$ \ref{eq:ce} \\
update $F, S$ using Equation \ref{eq:lms} \\
\textbf{Policy Generation} \\
update transfer path $h$ according to action $a$ \\
compute one-step reward using Equation \ref{eq:reward} \\
}
\textbf{Cumulative Reward} \\
compute cumulative reward using Equation \ref{eq:cumulative_reward} \\
}
\textbf{Policy Gradient Ascent} \\
compute the mean of the above cumulative reward \\
update $P$ using Equation \ref{eq:gradient} \\
}

\caption{Joint training algorithm}
\label{algr:training}
\end{algorithm}

\subsection{Policy Generation}
\label{sec:policy}


We employ Reinforcement Learning (RL) to generate policies for each intermediate domain. The components of this RL framework are defined as follows.

\textbf{State:} The state $s$ is defined by the domain-specific features. Specifically, $s$ at time $T$ is represented as $s_T=(\Phi_{T-1}^i,\Phi_T^i,\Phi_T^t)$, where $\Phi$ denotes the domain-specific features $f_{ds}$. Here, $\Phi_{T-1}^i$ corresponds to the domain-specific features of the previously selected intermediate domain, $\Phi_T^i$ to the current intermediate domain, and $\Phi_T^t$ to the target domain. The initial state, $\Phi_{T-1}^i$ is set to be the representation of the source domain, denoted as $\Phi_{T-1}^s$.

\textbf{Action:} The action \(a\) is a binary decision that can take values of either 0 or 1, which determines whether the current intermediate domain is selected during the policy generation process. The action $a$ is determined probabilistically to be 1 with probability $p$ and 0 with probability $1-p$, where $p$ is the output probability from the policy network. This binary decision constitutes the policy $\pi(a|s)$. After each action, the policy network updates the domain-specific feature representation $\Phi_{T}^i$, resulting in a transition of the state $s$ to ${s}^{\prime}$.


\textbf{Reward:} 
The reward \(R_T(s, a, s^{\prime})\) is generated through an unsupervised reward mechanism that measures the distance between domain-specific features across various domains. This reward is calculated based on future states and is defined as follows:


\begin{equation}
R_T= 
\begin{cases}
2\mathbb{D}_{it}^{s} - \mathbb{D}_{ii}^{s^{\prime}} - \mathbb{D}_{it}^{s^{\prime}} &  \text{if} \ a=1, \left( \mathbb{D}_{ii}^{s^{\prime}} < \mathbb{D}_{it}^{s} \right) \text{and} \left(  \mathbb{D}_{it}^{s^{\prime}} < \mathbb{D}_{it}^{s} \right)
 \\
-inf &  \text{if} \ a=1, \left( \mathbb{D}_{ii}^{s^{\prime}} \geq \mathbb{D}_{it}^{s} \right) \text{or} \left(  \mathbb{D}_{it}^{s^{\prime}} \geq \mathbb{D}_{it}^{s} \right)
\\
0 &  \text{if} \ a=0.
\end{cases}
\label{eq:reward}
\end{equation}
where $s^{\prime}$ denotes the subsequent state. The terms $\mathbb{D}_{it}$ and $\mathbb{D}_{ii}$ represent the distances $d(\Phi_T^i, \Phi_T^t)$ and $d(\Phi_T^i, \Phi_{T-1}^i)$, respectively, which measure the distance between the current intermediate domain and the target domain, and the distance between the current and previously selected intermediate domains. This reward strategy prioritizes selecting domains that shorten the transfer distance, specifically those where the distances to both the previously selected intermediate domain and the target domain are smaller than the distance between the previously selected intermediate domain and the target domain. The distance function $d$ is defined using the Wasserstein distance \cite{villani2009wasserstein}. This metric is essential for establishing generalization bounds in domain adaptation and is highly effective in reducing domain shift within the Wasserstein framework \cite{shen2018wasserstein, redko2020survey, liu2024enhancing}.

In this study, we employ a neural network-based policy generator within the RL framework, utilizing the classical policy gradient method \cite{silver2014deterministic}. This method models the policy parametrically as $\pi(a|s; \theta)$, optimized through gradient ascent to maximize the expected value of the value function, \(J(\theta) = \mathbb{E}_s[V(s; \theta)]\). 
The state value function $V(s;\theta)$ and the cumulative reward function \(Q(s, a)\) are defined as follows:
\begin{equation}
    \begin{gathered}
        V(s;\theta)=\sum_{a} \pi(a|s;\theta) \cdot Q(s, a), \\
        Q(s, a)=\mathbb{E}_{a\sim\pi(\cdot|s;\theta)}\left[\sum_{k=T}^{\infty} \gamma^{k-T} R_k(s, a, s^{\prime})\right],
    \end{gathered}
    \label{eq:cumulative_reward}
\end{equation}
where $\gamma$ denotes the discount factor, and $R(s, a, s^{\prime})$ represents the single-step reward as Equation \ref{eq:reward}. Policy updates are conducted using gradient ascent:

\begin{equation}
\begin{gathered}
    \theta \leftarrow \theta+\tau \nabla_{\theta} J(\theta), \\
    \nabla_\theta J(\theta)=\mathbb{E}_{a\sim\pi(\cdot|s;\theta)}[\nabla_\theta \log\pi(a|s;\theta) \cdot Q(s, a)],
    \label{eq:gradient_flow}
    \end{gathered}
\end{equation}
which adjusts the likelihood of actions based on their reward. By incorporating Equation \ref{eq:reward} and Equation \ref{eq:cumulative_reward} into Equation \ref{eq:gradient_flow}, we can derive the specific form of $\nabla_\theta J(\theta)$,

\begin{equation}
\begin{split}
\nabla_\theta J(\theta)=&a \nabla_\theta \log p \sum_{k=T}^{\infty} \gamma^{k-T} R_k(s, a, s^{\prime}) + \\
&(1-a) \nabla_\theta \log (1-p) \sum_{k=T}^{\infty} \gamma^{k-T} R_k(s, a, s^{\prime})
\label{eq:gradient}
\end{split}
\end{equation}

The policy gradient method leverages the advantages of reinforcement learning, allowing for dynamic adjustments to the policy in response to changes in domain-specific features. We adopt a joint training approach for both the feature extractor and the policy generator, as outlined in Algorithm \ref{algr:training}, to achieve simultaneous optimization of the transfer path and prediction outcomes.

\section{Experiment Results}
\subsection{Data Description}

We utilize two datasets in our study. The Rotated MNIST dataset \cite{nishida2023zero} expands upon the original MNIST by featuring images of digits rotated between 0 to 180 degrees, organized into 11 domains, with each domain containing 1,000 samples. The Alzheimer's Disease Neuroimaging Initiative (ADNI) dataset \cite{petersen2010alzheimer} provides MRI images to support Alzheimer's research. It comprises a source domain with individuals aged 50-70, including 190 samples, and several intermediate domains for ages 70-92, each with 50 samples, which are utilized for classifying into five disease categories.


\begin{table}[htbp]
\caption{\textbf{Accuracy for two datasets of different algorithms}}
\begin{center}
\begin{tabular}{c|l|cc}
 \toprule
\textbf{Category} & \textbf{\textbf{Method}} &  \textbf{ROT MNIST~($\uparrow$)} & \textbf{ADNI~($\uparrow$)}  \\

 \midrule
 \multirow{7}{*}{\textbf{CDA}} &  EAML \cite{liu2020learning}&70.4 &68.3 \\
  &  AGST \cite{zhou2022active} &76.2 & 57.3\\
  &  Gradual ST \cite{kumar2020understanding} &87.9 & 64.5\\
  &  CDOT \cite{ortiz2019cdot} &75.6 & 82.6\\
   & CMA \cite{hoffman2014continuous} &65.3 &55.4 \\
  &  W-MPOT \cite{liu2024enhancing} &89.1 & 88.3\\
  &  CIDA \cite{wang2020continuously} &85.7 & 73.6\\

 \midrule

 \multirow{5}{*}{\textbf{\makecell{CDA \\with RL}}}  & SDG \cite{liu2019reinforced} &89.8 & 80.5\\
  &  GRADIENT \cite{huang2022curriculum} &90.2 &88.4 \\
  &  CLEAS \cite{gao2022efficient} &\underline{92.8} & {86.1}\\
  & RCL \cite{xu2018reinforced} & 91.5& \underline{89.3}\\

&\textbf{Ours} &\textbf{93.4} & \textbf{90.5}\\
\bottomrule
\end{tabular}
\label{table:prediction}
\end{center}
\end{table}

\begin{figure}[ht]
	\begin{center}
		\centerline{\includegraphics[width=0.8\columnwidth]{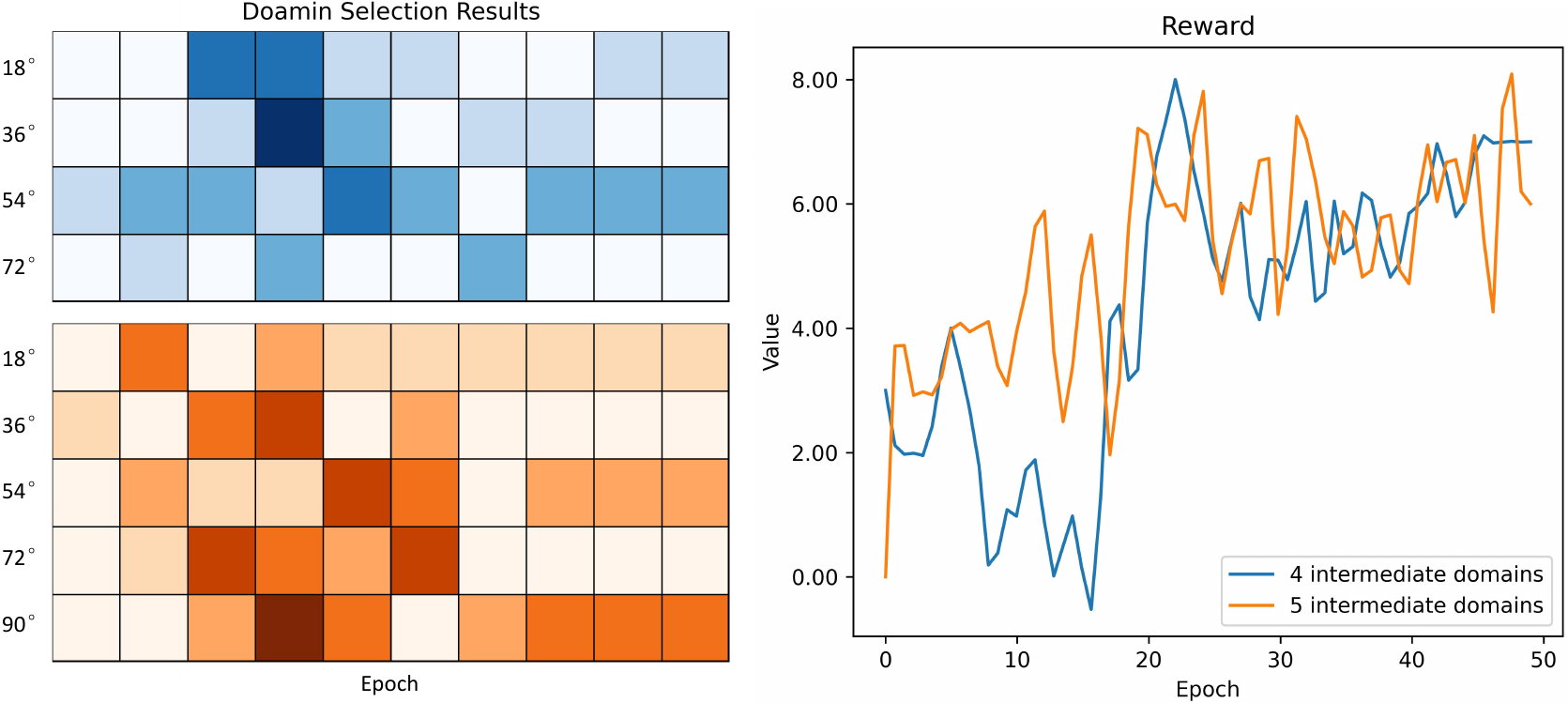}}
		\caption{\textbf{Reinforced Domain selection results.} The y-axis of the left figure represents various intermediate domains, each identified by a specific rotation angle ranging from the source domain at 0 degrees to the target domain, which is the last intermediate domain incremented by an additional 18 degrees. The x-axis samples every five epochs. A color gradient from light to dark illustrates the order of domain selection throughout the experiment, with white denoting domains that were not selected. The blue and yellow curves in the right figure represent setups with four and five intermediate domains, respectively, consistent with the configurations shown in the left figure.}
		\label{fig:reward}
	\end{center}
 \vspace{-0.5cm}
\end{figure}

\subsection{Quantitative and Qualitative Results}
We have compared classical Continuous Domain Adaptation (CDA) methods with those that incorporate Reinforcement Learning (RL) to address challenges within CDA, setting the number of intermediate domains at four. As demonstrated in Table \ref{table:prediction}, our approach achieves the highest accuracy on both datasets, with improvements of 0.6\% and 1.2\% on Rotated MNIST and ADNI, respectively. The greater gains on the ADNI dataset, which is higher-dimensional and more complex than MNIST, underscore our method's capability to effectively select paths and enhance performance in complex settings. Furthermore, the results demonstrate that RL significantly enhances CDA effectiveness, as evident from the superior performance in the lower half of the table.

\begin{figure}[ht]
	\begin{center}
		\centerline{\includegraphics[width=0.8\columnwidth]{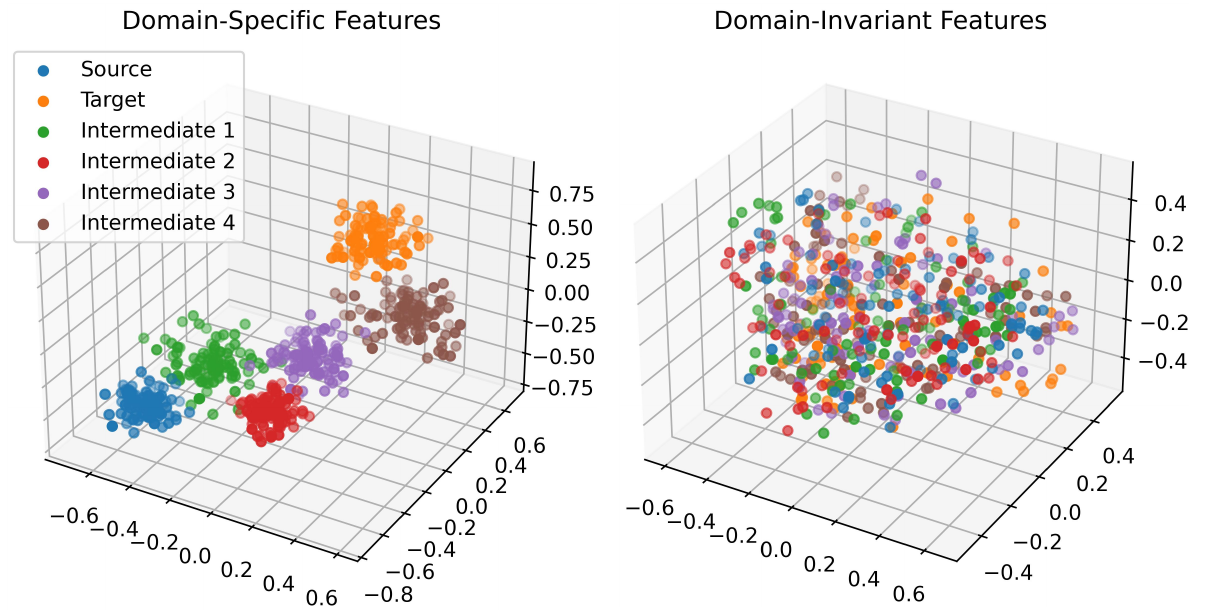}}
		\caption{\textbf{Visualizations of the domain-specific and domain-invariant features.} This image was generated using the t-SNE \cite{sejourne2021large} method and visualized by reducing the dimensions to three. Different colors represent different domains.}
		\label{fig:feature}
	\end{center}
 \vspace{-0.5cm}
\end{figure}

\begin{wraptable}{r}{3cm}
\caption{\textbf{Ablation Study for the number of intermediate domain pool}}
\begin{center}
\begin{tabular}{c|cc}
 \toprule
\textbf{K} &  \textbf{Accuracy}  \\
\midrule
2 & 92.8\\
3 & 93.2\\
4 & 93.4\\
5 & 93.8\\
6 & 94.1\\
7 & 93.5\\
8 & 93.2\\
\bottomrule
\end{tabular}
\label{table:inter}
\end{center}
\end{wraptable}
Figure \ref{fig:reward} demonstrates that as rewards increase, domain selection becomes more strategic, evidenced by a systematic increase in the angles of the selected domains. Upon reward stabilization, the model selectively omits intermediate domains with minor shifts or redundant details to minimize errors. The optimal paths for configurations with four and five intermediate domains are respectively $\hat{h}=(18, 54)$ and $\hat{h}=(18, 54, 90)$. Subsequently, Figure \ref{fig:feature} shows that the specific network effectively captures distinct features from each domain, which exhibit a continuity that facilitates continual domain transfer. Conversely, the invariant network extracts domain-invariant features, facilitating a unified feature distribution across various domains. This highlights the effectiveness of mutual information in ensuring consistent feature representation, regardless of domain variations.

\begin{table}[htbp]
\caption{\textbf{Ablation Study for the model Structure}}
\begin{center}
\begin{tabular}{ccc|c}
 \toprule
\textbf{\makecell{Classification\\ Model}} & \textbf{\makecell{Feature\\ Disentanglement}} & \textbf{\makecell{Policy \\ Generation}} & \textbf{Accuracy}  \\
\midrule
\cmark & \xmark&\xmark & 50.8\\
\cmark & \cmark &\xmark & 81.5\\
\cmark & \cmark & \cmark & \textbf{93.4}\\

\bottomrule 
\end{tabular}
\label{table:model}
\end{center}
\end{table}

\subsection{Ablation Study}

We conducted an ablation study on the Rotated MNIST dataset to evaluate the impact of feature disentanglement and policy generation within our CDA framework. As depicted in Table \ref{table:model}, incorporating these components increased the prediction accuracy to 93.4\%, highlighting their critical role in improving model performance. Additionally, we investigated the impact of varying the number of intermediate domain pool, denoted by $K$ (Table \ref{table:inter}). Contrary to typical CDA challenges, our method, enhanced by the policy generation mechanism, maintained high accuracy even with an increased number of intermediate domains, demonstrating its robust ability to manage domain transfer effectively.

\section{Conclusion}
Our study addresses the challenge of dynamic domain selection in Continuous Domain Adaptation by joint Reinforcement Learning and feature disentanglement, simultaneously optimizing the transfer path and improving prediction outcomes. Our domain selection policy, driven by an unsupervised reward mechanism based on distances between latent domain embeddings and learned through a policy gradient algorithm, significantly enhances strategic precision. By distinguishing domain-specific from domain-invariant features, our approach improves the extraction of transferable features vital for effective domain adaptation and enables more precise estimations of domain shifts using low-dimensional embeddings. Empirical validation on datasets like Rotated MNIST and ADNI confirms our method's superiority, surpassing traditional CDA approaches with improved prediction accuracy and more efficient path selection.

\section*{Acknowledgements}
This work is supported in part by the Natural Science Foundation of China (Grant 62371270), Shenzhen Key Laboratory of Ubiquitous Data Enabling (No.ZDSYS20220527171406015), and Tsinghua Shenzhen International Graduate School-Shenzhen Pengrui Endowed Professorship Scheme of Shenzhen Pengrui Foundation.

\bibliographystyle{IEEEtran}
\bibliography{mybibfile}

\end{document}